\documentclass{article} %
\usepackage{iclr2020_conference,times}
 
\usepackage{amsmath,amsfonts,bm}

\def\eqref#1{equation~\ref{#1}}

\def\1{\bm{1}}

\DeclareMathAlphabet{\mathsfit}{\encodingdefault}{\sfdefault}{m}{sl}
\SetMathAlphabet{\mathsfit}{bold}{\encodingdefault}{\sfdefault}{bx}{n}

\usepackage[utf8]{inputenc} %
\usepackage[T1]{fontenc}    %
\usepackage{hyperref}       %
\usepackage{url}            %
\usepackage{booktabs}       %
\usepackage{amsfonts}       %
\usepackage{nicefrac}       %
\usepackage{microtype}      %
\usepackage{physics}        %
\usepackage{empheq}
\usepackage[most]{tcolorbox}
\newtcbox{\mymath}[1][]{%
    nobeforeafter,
    math upper, tcbox raise base,
    enhanced,
    colframe=white!10!black,
    colback=white!10,
    boxrule=1pt,
    #1} %
\tcbset{highlight math style={boxsep=1mm,colback=blue!10!red!10!white}}

\usepackage{amsmath}
\usepackage{amssymb}
\usepackage{graphicx}
\usepackage{dsfont}
\usepackage{booktabs}
\usepackage{multirow}
\usepackage{wrapfig}
\usepackage{float}
\usepackage{subfigure}
\usepackage{caption}
\usepackage{ownstyles}
\usepackage{tabularx}
\usepackage{chngcntr}
\usepackage{cleveref}
\usepackage{xspace}

\title{Lagrangian Neural Networks}

\newif\ifarxiv
\arxivtrue
 
\author{Miles Cranmer \\
Princeton University \\
\texttt{mcranmer} \\
\texttt{@princeton.edu}
\And
Sam Greydanus \\
Oregon State University \\
\texttt{greydanus.17} \\
\texttt{@gmail.com}
\And
Stephan Hoyer \\
Google Research \\
\texttt{shoyer} \\
\texttt{@google.com}
\And
Peter Battaglia \\
DeepMind \\
\texttt{peterbattaglia} \\
\texttt{@google.com}
\And
David Spergel\thanks{Also affiliated with Princeton University}\\
Flatiron Institute \\
\texttt{davidspergel} \\
\texttt{@flatironinstitute.org}
\And
Shirley Ho\thanks{Also affiliated with New York University, Princeton University, and Carnegie Mellon University}\\
Flatiron Institute \\
\texttt{shirleyho} \\
\texttt{@flatironinstitute.org}
}

\newcommand\jax{\textsc{JAX}\xspace}
 
\renewcommand{\d}{\dot}
\renewcommand{\dd}{\ddot}
\newcommand{\df}[2]{\frac{\partial #1}{\partial #2}}

\iclrfinalcopy %
\newcommand\lnn{LNN\xspace}
\newcommand\lnns{LNNs\xspace}

\newcommand\fulllgn{Lagrangian Graph Network\xspace}
\newcommand\fulllgns{Lagrangian Graph Networks\xspace}

\begin{document}

\maketitle
 
\begin{abstract}
 
Accurate models of the world are built upon notions of its underlying symmetries. In physics, these symmetries correspond to conservation laws, such as for energy and momentum. Yet even though neural network models see increasing use in the physical sciences, they struggle to learn these symmetries. In this paper, we propose Lagrangian Neural Networks (\lnns), which can parameterize arbitrary Lagrangians using neural networks. In contrast to models that learn Hamiltonians, \lnns do not require canonical coordinates, and thus perform well in situations where canonical momenta are unknown or difficult to compute. Unlike previous approaches, our method does not restrict the functional form of learned energies and will produce energy-conserving models for a variety of tasks. We test our approach on a double pendulum and a relativistic particle, demonstrating energy conservation where a baseline approach incurs dissipation and modeling relativity without canonical coordinates where a Hamiltonian approach fails.
\ifarxiv
Finally, we show how this model
can be applied to
graphs and continuous systems using a \fulllgn, and demonstrate it on
the 1D wave equation.
\fi
\end{abstract}

\begin{figure}[h!]
\centering
    \includegraphics[width=.9\textwidth]{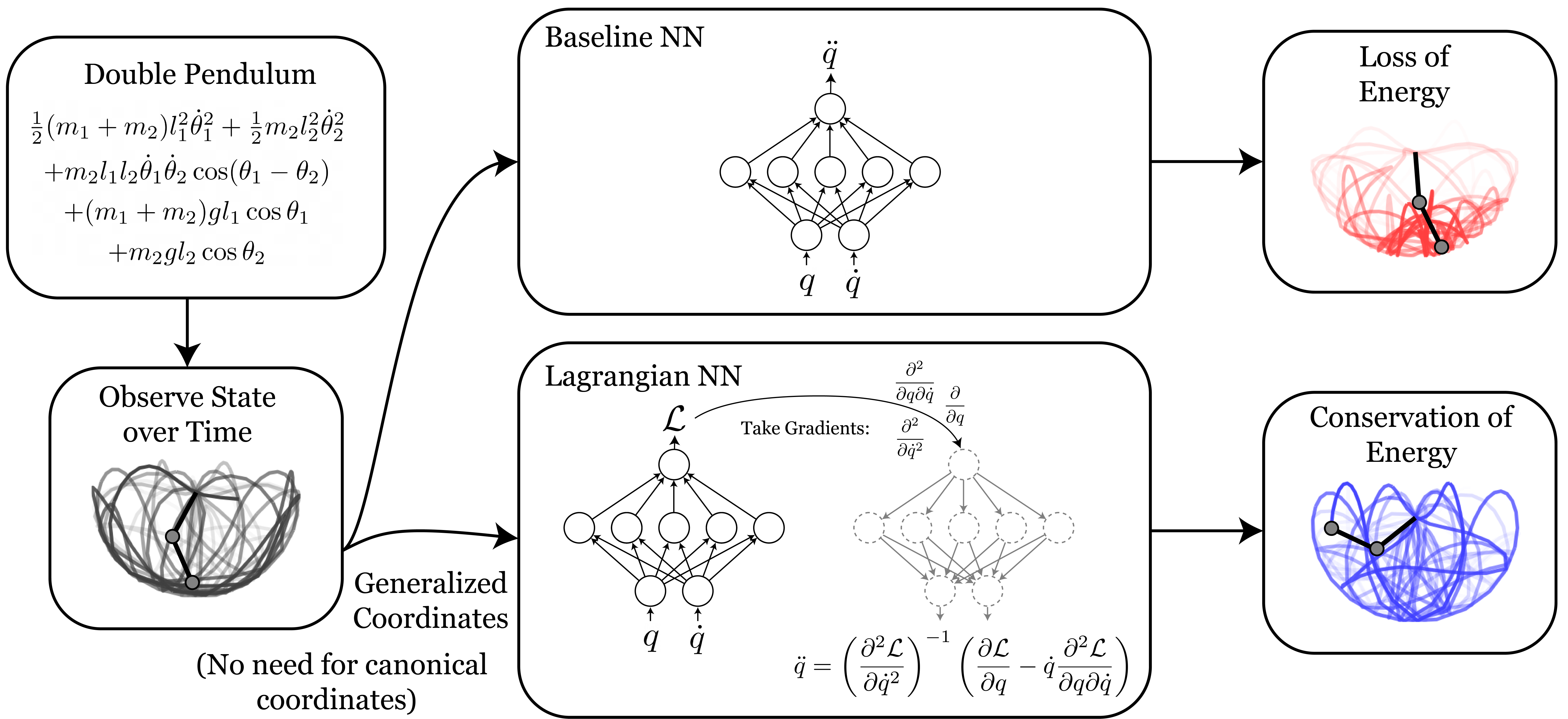}%
    \caption{
    Physicists use Lagrangians to describe the dynamics of physical systems like the double pendulum (black). Neural networks struggle to model these dynamics over long time periods due to their inability to conserve energy (red). In this paper, we show how to learn arbitrary Lagrangians with neural networks, inducing a strong physical prior on the learned dynamics (blue).}
 
\figlabel{fig1}
\end{figure}
 
\section{Introduction}

Neural networks excel at finding patterns in noisy and high-dimensional data.
They can perform well on tasks such as image classification \citep{krizhevsky2012imagenet-classification-with-deep}, language translation \citep{sutskever-2014-arXiv-sequence-to-sequence-learning}, and game playing \citep{silver2017mastering}. Part of their success is rooted in "deep priors" which include hard-coded translation invariances (e.g., convolutional filters), clever architectural choices (e.g., self-attention layers), and well-conditioned optimization landscapes (e.g., batch normalization).
Yet, in spite of their ability to compete with
humans on challenging tasks, these models lack many basic
intuitions about the dynamics of the physical world. Whereas
a human can quickly deduce that a ball thrown upward will
return to their hand at roughly the same velocity, a neural
network may never grasp this abstraction, even after seeing thousands of examples \citep{bakhtin2019phyre}.
 
The basic problem with neural network models is that they struggle to learn basic symmetries and conservation laws. One solution to this problem is to design neural networks that can learn arbitrary conservation laws. This was the core motivation behind Hamiltonian Neural Networks by \cite{greydanus} and Hamiltonian Generative Networks by \cite{toth2019hamiltonian}. These are both types of differential equations modeled by neural networks, or ``Neural ODEs,'' which were studied extensively in \cite{ricky}.
 
Yet while these models were able to learn effective conservation laws, the Hamiltonian formalism requires that the coordinates of the system be ``canonical.'' In order to be canonical, the input coordinates $(q, p)$
to the model should obey a strict set of rules given by the Poisson bracket relations:
\begin{align}
p_i \equiv \frac{\partial \mathcal{L}}{\partial \dot{q}_i}
\Longleftrightarrow
    \{q_i,q_j\} &= 0 \quad \{p_i,p_j\} = 0 \quad \{q_i,p_j\} = \delta_{ij},\\
    \text{ where }
    \{ f, g\} &= 
    \sum_i 
    \left( 
    \df{f}{q_i}
    \df{g}{p_i}
    -
    \df{f}{p_i}
    \df{g}{q_i}
    \right), 
    \nonumber
\end{align}
where $\dot{q}$ indicates a time derivative and $\mathcal{L}$ the Lagrangian. The problem is that many datasets have dynamical coordinates that do not satisfy this constraint: $p_i$ is not simply $\dot{q}_i$ $\times$ mass in many cases. A promising alternative is to learn the Lagrangian of the system instead. Like Hamiltonians, Lagrangians enforce conservation of total energy, but unlike Hamiltonians, they can do so using arbitrary coordinates.
 
In this paper, we show how to learn Lagrangians using neural networks. We demonstrate that they can model complex physical systems which Hamiltonian Neural Networks (HNNs) fail to model while outperforming a baseline neural network in energy conservation. We discuss our work in the context of ``Deep Lagrangian Networks,'' (DeLaNs) described in \cite{lutter2019delan}, a closely related method for learning Lagrangians. Unlike our work, DeLaNs are built for continuous control applications and only models rigid body dynamics. Our model is more general in that it does not restrict the functional form of the Lagrangian.
\ifarxiv
We also show how our model
can be applied to continuous systems
and graphs using a \fulllgn in \cref{sec:exp}.
\fi
 
\begin{table}[h!]
\caption{An overview of neural network based models for physical dynamics. Lagrangian Neural Networks combine many desirable properties. DeLaNs explicitly restrict the learned kinetic energy, and HNNs implicitly do as well by requiring a definition of the canonical momenta.}
\tablabel{tab1}
\begin{center}
\begin{tabular}{@{}lccccc@{}}
\toprule
 & Neural net & Neural ODE & HNN & DeLaN & \lnn  (ours) \\ \midrule
Can model dynamical systems & \checkmark & \checkmark & \checkmark & \checkmark & \checkmark  \\
Learns differential equations &  & \checkmark & \checkmark & \checkmark & \checkmark  \\
Learns exact conservation laws &  &  & \checkmark & \checkmark & \checkmark  \\
Learns from arbitrary coords.& \checkmark & \checkmark &  & \checkmark & \checkmark  \\ 
Learns arbitrary Lagrangians & & &  &  & \checkmark  \\  %
\ifarxiv
\fi
\bottomrule
\end{tabular}
\end{center}
\end{table}
 
\section{Theory}
 
\textbf{Lagrangians.} The Lagrangian formalism models a classical physics system with coordinates $x_t = (q, \dot q)$ that begin in one state $x_0$ and end up in another state $x_1$. There are many paths that these coordinates might take as they pass from $x_0$ to $x_1$, and we associate each of these paths with a scalar value called ``the action.'' Lagrangian mechanics tells us that if we let the action be the functional:
\begin{align}
S ~=~ \int_{t_0}^{t_1} \left( T(q_t, \dot q_t) - V(q_t)\right) ~~ dt,
\end{align}
where $T$ is kinetic energy and $V$ is potential energy, then there is only one path that the physical system will take, and that path is the stationary (e.g., minimum) value of $S$. In order to enforce the requirement that $S$ be stationary, i.e., $\delta S=0$, we define the Lagrangian to be $\mathcal{L}\equiv T-V$, and derive a constraint equation called the \textit{Euler-Lagrange equation} which describes the path of the system: 
\begin{align}
\frac{d}{dt} \frac{\partial \mathcal{L}}{\partial \dot q_j} = \frac{\partial \mathcal{L}}{\partial q_j}.
\end{align}
 
\textbf{Euler-Lagrange with a parametric Lagrangian.} Physicists traditionally write down an analytical expression for $\mathcal{L}$ and then symbolically expand the Euler-Lagrange equation into a system of differential equations. However, since $\mathcal{L}$ is now a black box, we must resign ourselves to the fact that there can be no analytical expansion of the Euler-Lagrange equation. Fortunately, we can still derive a numerical expression for the dynamics of the system. We begin by rewriting the Euler-Lagrange equation in vectorized form as
\begin{equation}
    \frac{d}{dt} \nabla_{\dot q} \mathcal{L} = \nabla_{q} \mathcal{L}
\end{equation}
where $\left(\nabla_{\dot{q}}\right)_i\equiv \frac{\partial}{\partial \d{q}_i}$. Then we can use the chain rule to expand the time derivative $\frac{d}{dt}$ through the gradient of the Lagrangian, giving us two new terms, one with a $\ddot q$ and another with a $\dot q$:
    \begin{equation}
(\nabla_{\dot q}\nabla_{\dot q}^{\top}\mathcal{L})\ddot q + (\nabla_{q}\nabla_{\dot q}^{\top}\mathcal{L}) \dot q = \nabla_q \mathcal{L}.
    \end{equation}
Here, these products of $\nabla$ operators are matrices, e.g.,
$(\nabla_{q}\nabla_{\dot q}^{\top}\mathcal{L})_{ij} = \frac{\partial^2 \mathcal{L}}{\partial q_j \partial \dot{q}_i }$.
Now we can use a matrix inverse to solve for $\ddot q$:
\begin{empheq}[box=\mymath]{equation}
    \dd{q} = (\nabla_{\dot q}\nabla_{\dot q}^{\top}\mathcal{L})^{-1}[\nabla_q \mathcal{L} - (\nabla_{q}\nabla_{\dot q}^{\top}\mathcal{L})\dot q].
    \label{eqn:blackbox-el}
\end{empheq}
For a given set of coordinates $x_t=(q_t,\dot q_t)$, we now have a method for calculating $\ddot q_t$ from a black box Lagrangian, which we can integrate to find the dynamics of the system. In the same manner as \cite{greydanus}, we can also write a loss function in terms
of the discrepancy between $\ddot x_t^{\mathcal{L}}$ and $\ddot x_t^{\textrm{true}}$.

\renewcommand\L{\mathcal{L}}
\section{Related Work}
 
\textbf{Physics priors.} A variety of previous works have sought to endow neural networks with physics-motivated inductive biases. These include work for domain-specific problems in molecular dynamics \citep{rupp2012fast}, quantum mechanics \citep{schutt2017quantum}, or robotics \citep{gupta,lutter2019delan}. Other approaches are more general, such as Interaction Networks \citep{battaglia2016interaction}, which models physical interactions as message passing on a graph.
 
\textbf{Learning invariant quantities.} Recent work by \cite{greydanus}, \cite{toth2019hamiltonian}, and \cite{chen2019symplectic} built on previous approaches of endowing neural networks with physical priors by demonstrating how to learn invariant quantities by approximating a Hamiltonian with a neural network. In this paper, we follow the same approach as \cite{greydanus}, but with the objective of learning a Lagrangian rather than a Hamiltonian so not to restrict the learned kinetic energy.
 
\textbf{DeLaN.} A closely related previous work is ``Deep Lagrangian Networks'', or DeLaN \citep{lutter2019delan}, in which the authors show how to learn specific types of Lagrangian systems. They assume that the kinetic energy is an inner product of the velocity: $T = \dot{q}^T M \dot{q}$, where $M$ is a $q$-dependent
positive definite matrix.
This approach works well for rigid body dynamics, which includes many systems encountered in robotics.
However, many systems do not have this kinetic energy, including, for example,
a charged particle in a magnetic field, and a fast-moving object in special relativity.
Other Lagrangian-based approaches include \cite{gupta, qin}.

\section{Methods}
 
\textbf{Solving Euler-Lagrange with \jax.}
Efficiently implementing \eqref{eqn:blackbox-el} represents a formidable technical challenge. In order to train this forward model, we need to compute the inverse Hessian $(\nabla_{\dot q}\nabla_{\dot q}^{\top}\mathcal{L})^{-1}$ (we use the pseudoinverse to avoid potential singular matrices) of a neural network and then perform backpropagation.
\ifarxiv
The matrix inverse scales as $\mathcal{O}(d^3)$ with the number of coordinates $d$.
\fi
Perhaps surprisingly, though, we can implement this operation in a few lines of \jax \citep{citejax} (see Appendix). We publish our code on GitHub\footnote{
\ifarxiv
\url{https://github.com/MilesCranmer/lagrangian_nns}
\else
Anonymized.
\fi
}.

\textbf{Training details.} For both baseline, HNN, and \lnn models, we used a four-layer neural network model with 500 hidden units,
a decaying learning rate starting at $10^{-3}$, and a batch size of 32.
We initialize our model using a novel initialization scheme described
in \cref{sec:init}, which was optimized for \lnns. 
 
\textbf{Activation functions.} Since we take the Hessian of a \lnn, the second-order derivative of the activation function is important. For example, the ReLU nonlinearity is a poor choice because its second-order derivative is zero. In order to find a better activation function, we performed a hyperparameter search over ReLU$^2$ (squared), ReLU$^3$, tanh, sigmoid, and softplus activations on the double pendulum problem. We found that softplus performed best and thus used it for all experiments.

\section{Experiments}
\label{sec:exp}
 
\textbf{Double pendulum}. The first task we considered was a dataset of simulated double pendulum trajectories. We set the masses and lengths to $1$ and learned the instantaneous accelerations over 600,000 random initial conditions. We found that the \lnn and the baseline converged to similar final losses of $7.3$ and $7.4 \times 10^{-2}$, respectively. The more significant difference was in energy conservation; the \lnn almost exactly conserved the true energy over time, whereas the baseline did not. Averaging over 40 random initial conditions with 100 time steps each, the mean energy discrepancy between the true total energy and predicted was 8\% and 0.4\% of the max potential energy for the baseline and \lnn models respectively.

\begin{figure}[H]
\centering
\subfigure[Error in angle]{\figlabel{fig1a}\includegraphics[width=.49\textwidth]{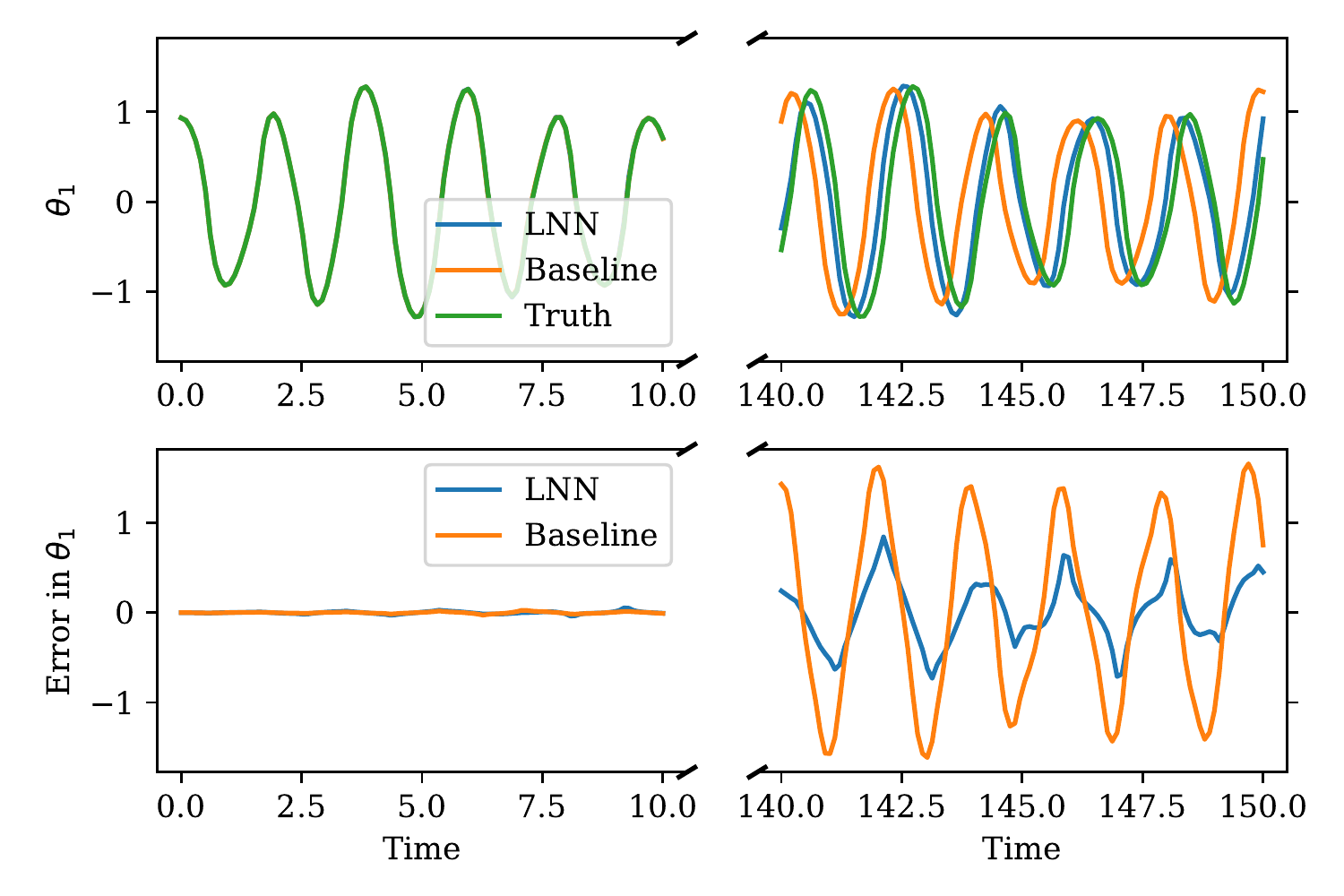}}
\subfigure[Error in energy]{\figlabel{appendix:fig1b}\includegraphics[width=.49\textwidth]{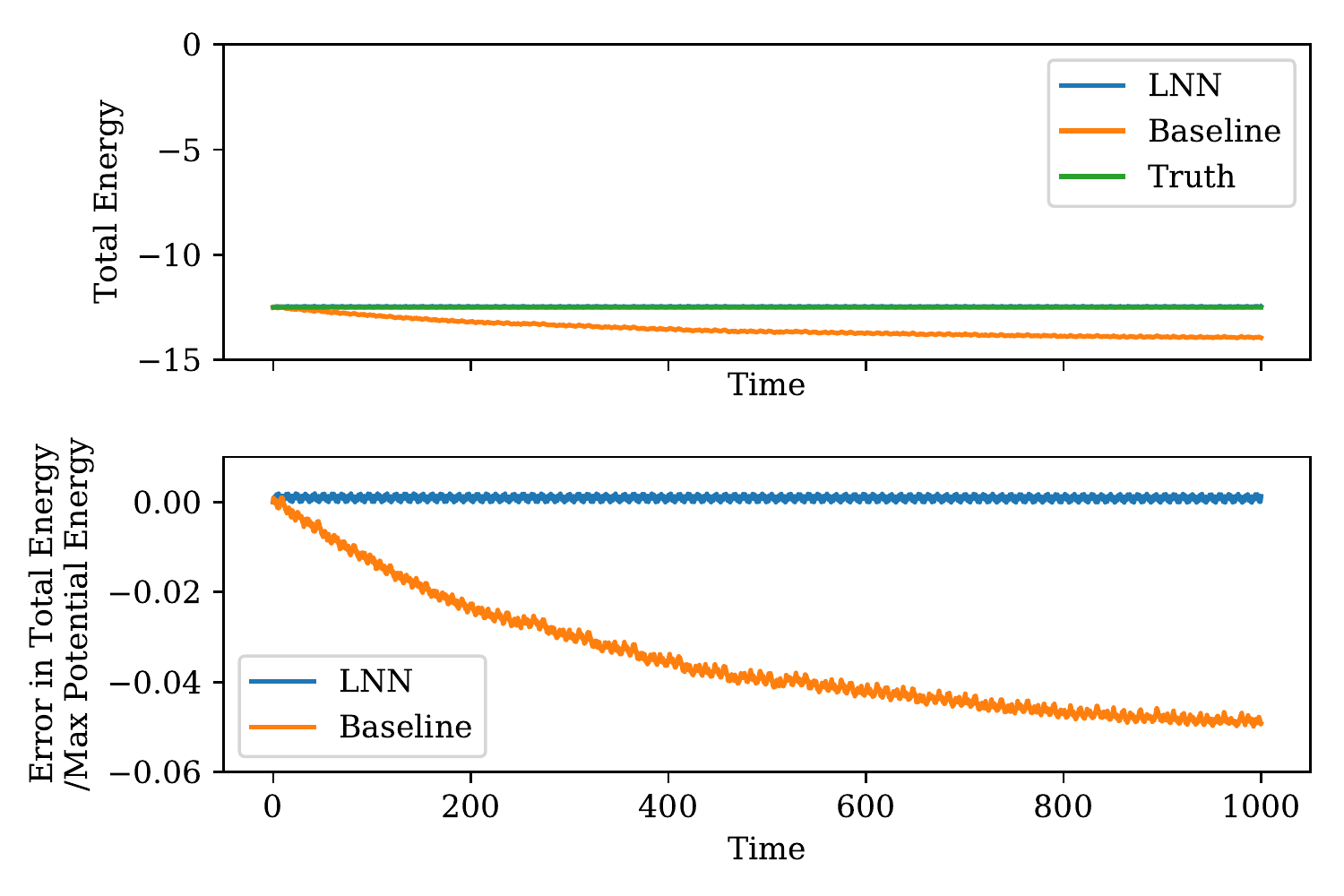}}
\caption{Results on the double pendulum task. In (a), we see that
the \lnn and baseline model perform similarly in modelling the dynamics of the pendulum over short time periods. However, if we plot out the energy over a very long time period in (b) we can see that the \lnn model conserves the total energy of the system significantly better than the baseline.}
\label{fig:dblpend-loss}
\end{figure}
 
\begin{figure}[H]
\centering
\subfigure[HNN, arbitrary coords.]{\includegraphics[width=.32\textwidth]{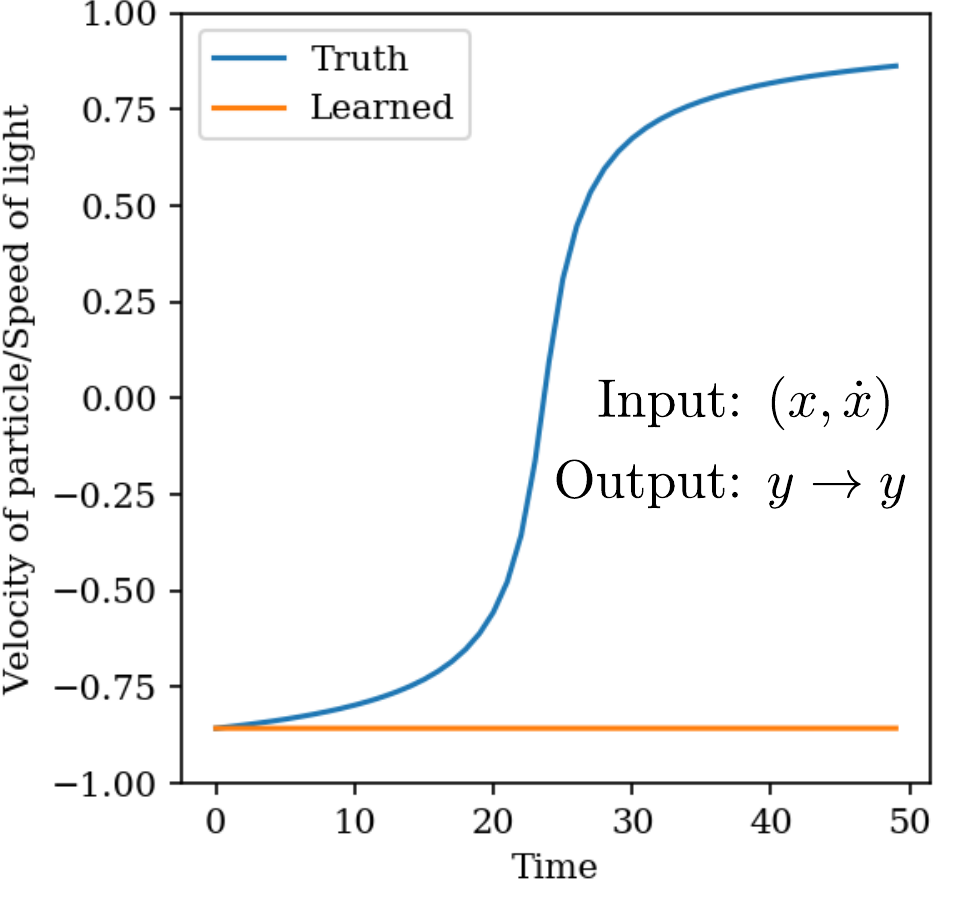}}
\subfigure[HNN, canonical coords.]{\includegraphics[width=.32\textwidth]{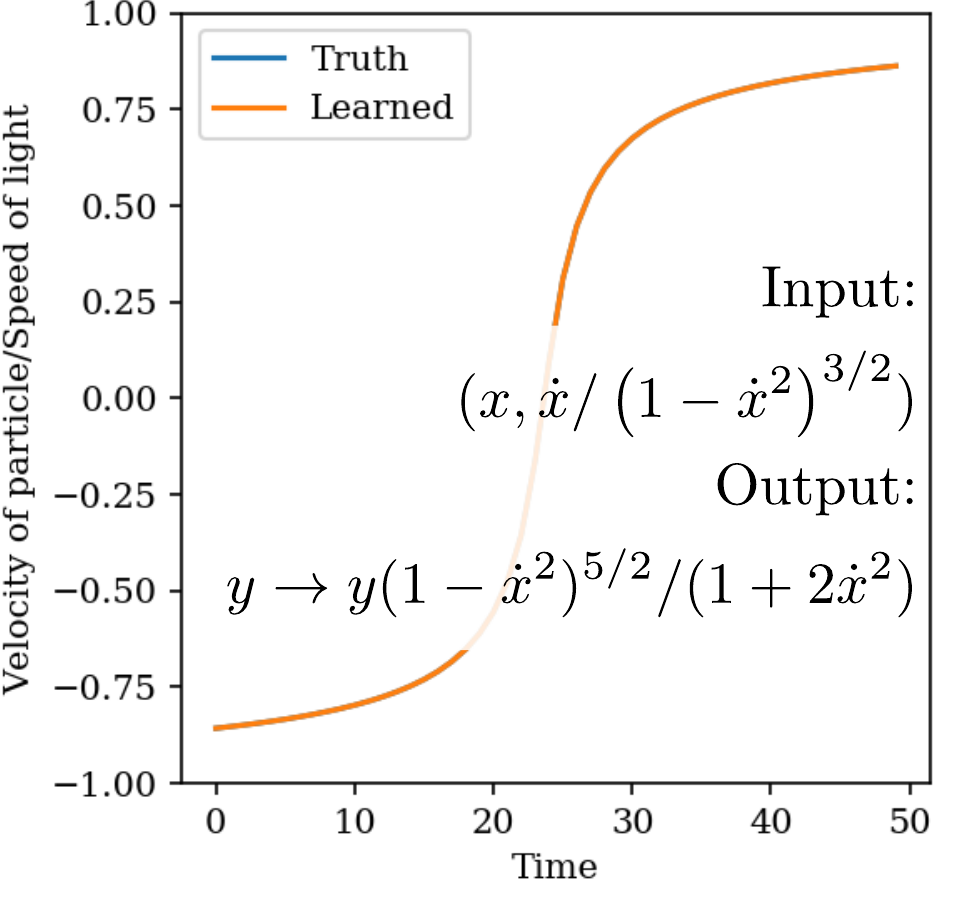}}
\subfigure[LNN, arbitrary coords.]{\includegraphics[width=.32\textwidth]{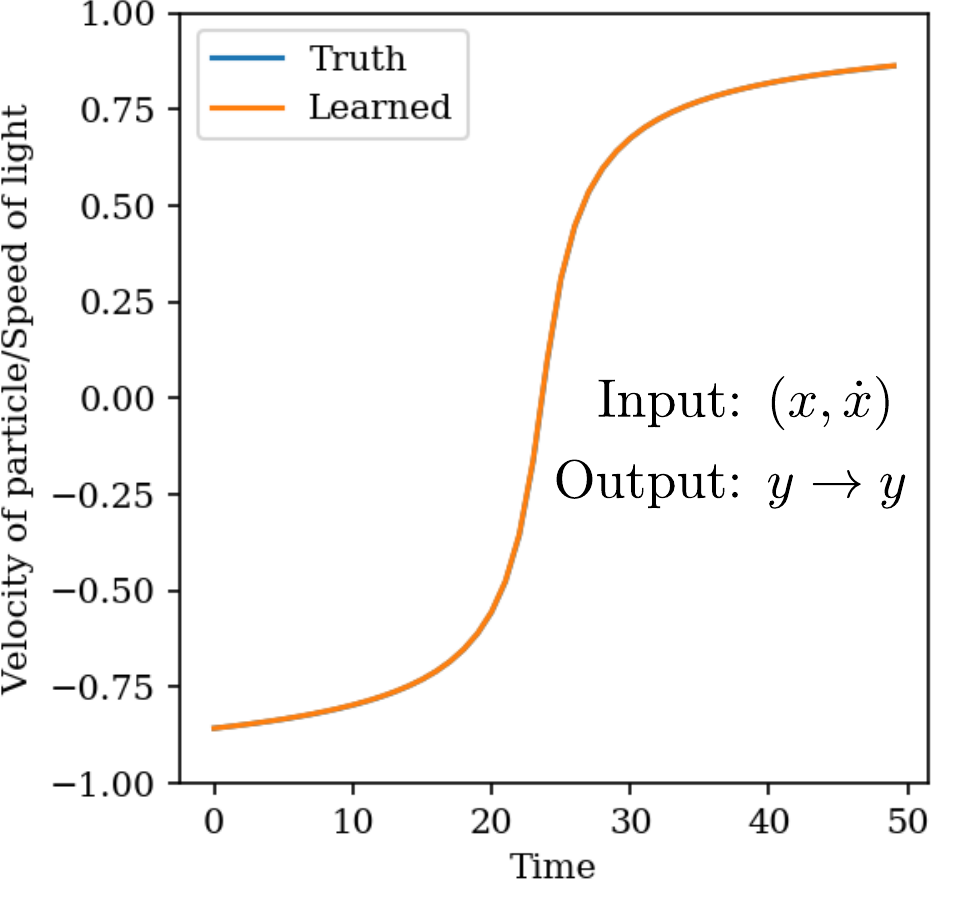}}
\caption{Results on the relativistic particle task. In (a) an HNN model fails to learn dynamics from non-canonical coordinates. In (b) the HNN succeeds when given canonical coordinates. Finally in (c) the LNN learns accurate dynamics even from non-canonical coordinates.}
\label{fig:sr}
\end{figure}

\textbf{Relativistic particle in a uniform potential}.
The second task was a relativistic particle in a uniform potential. For a particle with a mass of $1$ in a potential $g$ and with $c=1$, special relativity gives the Lagrangian ${\mathcal{L} = ((1 - \dot{q}^2)^{-1/2} - 1) + g q}$. The canonical momenta of this system are $\dot{q}(1 - \dot{q}^2)^{-3/2}$, which means that a Hamiltonian Neural Network will fail if given simple observables like $\dot q$ and $q$. The DeLaN model will also struggle since it assumes that $T$ is second order in $\dot{q}$. To verify these predictions, we trained HNN and \lnn models on systems with random initial conditions and values of $g$. Figure \ref{fig:sr} shows that the HNN fails without canonical coordinates whereas the LNN can work without this extra \textit{a priori} knowledge, and learns the system as accurately as an HNN trained on canonical coordinates.
 
\ifarxiv
 
\textbf{Wave equation with \fulllgns.} 
\Cref{eqn:blackbox-el} is applicable to many different types of systems,
including those with disconnected coordinates
such as graph-like and grid-like systems.
To model this, we now learn a Lagrangian density
which is summed to form the full Lagrangian.
This is similar to the Hamiltonian Graph Network of \cite{alvaro},
or more specifically, the
Flattened Hamiltonian Graph Network of \cite{cranmersymb}.
For simplicity, we focus on 1D grids with
locally-dependent dynamics (i.e., nodes
are connected to their two adjacent nodes).
 
A continuous material can be described in terms of
quantities $\phi_i$, such as the displacement
of a guitar string, at each gridpoint $x_i$.
One way to model this type of system is to treat
the quantities on adjacent gridpoints as independent coordinates,
and sum the regular \lnn equation over all
connected groups of coordinates.
For $n$ gridpoints, the total Lagrangian
is then:
\newcommand\Ld{\L_\text{density}}
\newcommand{\indices}{}
\begin{align}
    \L &= \sum_{i=1:n}\L_i,
    \text{ for }\L_i =\Ld\left(\{\phi_j, \dot{\phi}_j\}_{j\in{\mathcal{I}}_i} \right)
\end{align}
where ${\mathcal{I}}_i=\{i, \ldots\}$
is the set of indices connected to $i$.
For a 1D
grid where only the adjacent gridpoints
affect the dynamics of the center gridpoint,
this is $\{i, i-1, i+1\}$.
Again, we can solve dynamics by plugging the Lagrangian into \cref{eqn:blackbox-el},
now with $\phi$ as the coordinates:
\begin{equation}
 {\dd{\phi}
     = 
     \left(
     \nabla_{\dot{\phi}_{\indices}}
  \nabla_{\dot{\phi}_{\indices}}^\top
     \L
     \right)^{-1}
     \left(
     \nabla_{\phi_{\indices}} \L -
     \left( \nabla_{\phi_{\indices}}
     \nabla_{\dot{\phi}_{\indices}}^\top
     \L
     \right)\dot{\phi}_{\indices}
     \right)},
\end{equation}
where, e.g., $\nabla_{\phi_{\indices}} \equiv \{\df{}{\phi_1}, \df{}{\phi_2}, \ldots, \df{}{\phi_n}\}$.
Note that the Hessian matrix is sparse with non-zero entries at ``neighbor of neighbor'' positions in the graph, which can make it much more efficient to calculate and invert.
For example, in 1D it can be calculated with only 5 forward over backwards auto-differentiation passes and can be inverted in linear time.

We can think of this as a regular \lnn
where,
instead of calculating the Lagrangian
directly from a fixed set of coordinates,
we are now accumulating
a Lagrangian density over groups of coordinates. 
For a different connectivity, such as a graph
network, or to approximate
higher-order spatial derivatives for a continuous
material,
one would select $\mathcal{I}_i$
based on the adjacency matrix for the graph.
This model is a type of
Graph Neural Network \citep{scarselli2008graph}.
The Lagrangian density itself, $\Ld$,
we model as an MLP.
Since we write our models in \jax, we can
easily vectorize this forward model over the grid.
 
We consider the 1D wave equation, with
$\phi$ representing the wave displacement: $\phi(x, t)$.
For wave speed $c=1$, the equation describing its dynamics
can be written as: $\ddot{\phi} = \frac{\partial^2 \phi}{\partial x^2}$.
The Lagrangian for this differential equation is:
$\mathcal{L} = \int\left(\dot{\phi}^2 - \left(\df{\phi}{x}\right)^2 \right)dx.$
For the \lnn to learn this, the MLP will 
need to learn to approximate
a finite difference operator:
$\L_i = \dot{\phi}_i^2 - \left(\frac{\phi_{i+1} - \phi_{i-1}}{2 \Delta x}\right)^2$
for grid spacing $\Delta x$ (or any translation + scaling
of this equation).
We simulate the wave equation in a box
with periodic boundary conditions, and
learn the dynamics with this \fulllgn,
shown in \cref{fig:cnn}. The \fulllgn
models the wave equation accurately
and almost exactly conserves energy
integrated across the material.
 
\begin{figure}[h]
    \centering
    \href{https://github.com/MilesCranmer/gifs/blob/master/wave_equation.gif}{\includegraphics[width=0.99\linewidth]{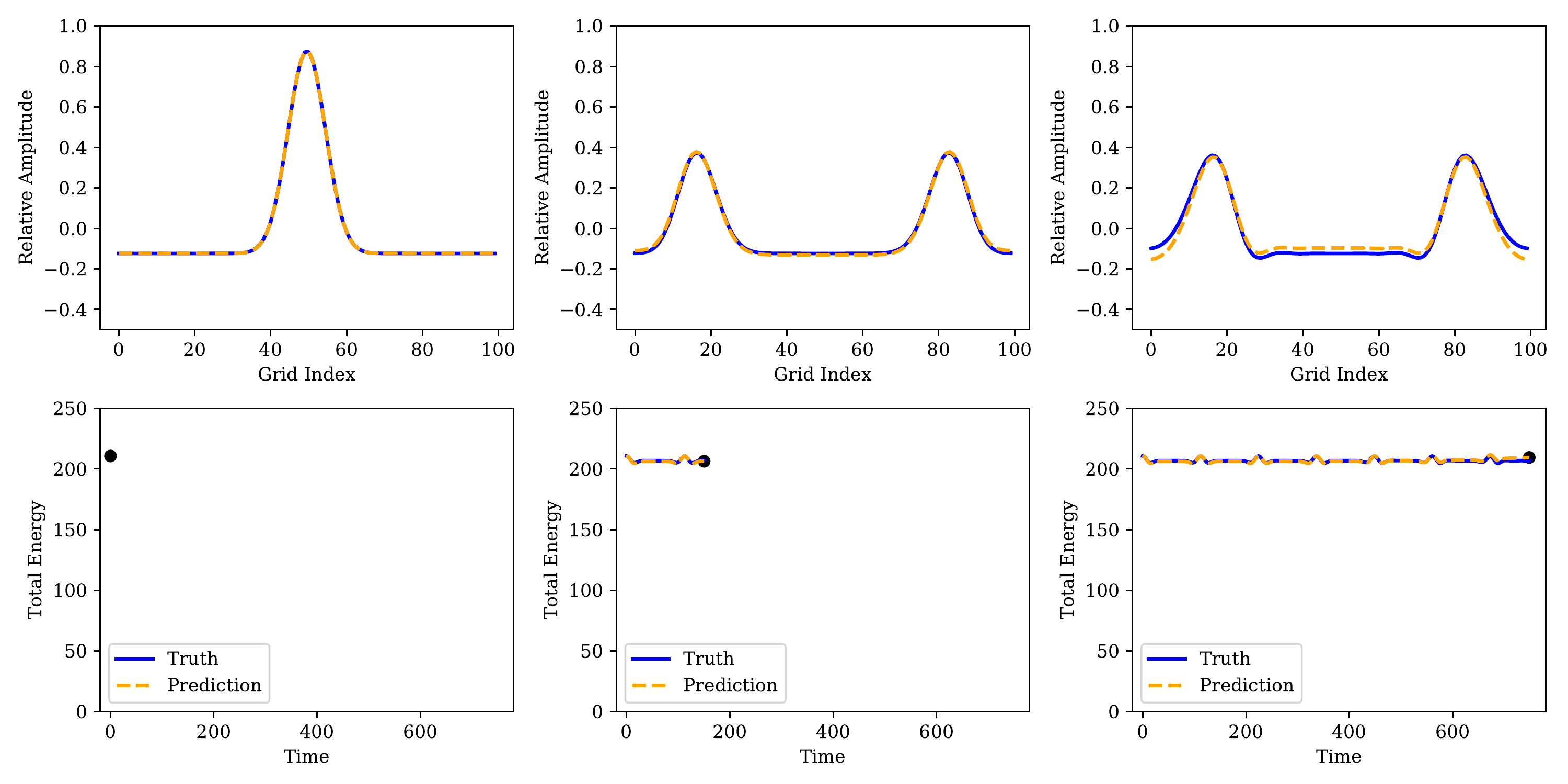}}
    \caption{Results on the 1D wave equation
    task, using a \fulllgn. Here,
    the \lnn models the Lagrangian density
    over three adjacent gridpoints along the wave,
    and this density is summed.
    Here, the waves make approximately three and a half
    passes across the 100 grid points
    over the time plotted. A movie can be
    viewed by clicking on the plot or by going to
    \url{https://github.com/MilesCranmer/gifs/blob/master/wave_equation.gif}.
    }%
    \label{fig:cnn}
\end{figure}
 
\fi
 
\section{Conclusion}
 
We have introduced a new class of neural networks, Lagrangian Neural Networks, which can learn arbitrary Lagrangians. In contrast to models that learn Hamiltonians, these models do not require canonical coordinates and thus perform well in situations where canonical momenta are unknown or difficult to compute. To evaluate our model, we showed that it could effectively conserve total energy on a complex physical system: the double pendulum. We showed that our model could learn non-trivial canonical momenta on a task where Hamiltonian learning struggles.
Finally, we demonstrated a graph version of the model with
the 1D wave equation.

\ifarxiv
\textbf{Acknowledgments}
Miles Cranmer would like to thank Jeremy Goodman for several discussions and validation of the theory behind the Lagrangian approach, and Adam Burrows and Oliver Philcox for very helpful comments on a draft of the paper.
\fi
 
\bibliography{main}
\bibliographystyle{iclr2020_conference}

\cleardoublepage
\appendix
{
\Large
\textbf{Appendix}
}
 
\section{Solving Euler-Lagrange with \jax} \label{sec:jax}
 
Here we will present a simple \jax implementation of Equation \ref{eqn:blackbox-el}. Assume that \texttt{lagrangian} is a differentiable
function that takes three vectors as input and outputs a scalar. Meanwhile, \texttt{q\_t} is a vector of the velocities ($\d{q}$) of \texttt{q} ($q$) and \texttt{q\_tt} contains the accelerations ($\dd{q}$). The vector \texttt{m} represents non-dynamical
parameters.

\begin{verbatim}
q_tt = (
 jax.numpy.linalg.pinv(jax.hessian(lagrangian, 1)(q, q_t, m)) @ (
    jax.grad(lagrangian, 0)(q, q_t, m)
  - jax.jacobian(jax.jacobian(lagrangian, 1), 0)(q, q_t, m) @ q_t
 )
)
\end{verbatim}
 
When this is called in a loss function with the \lnn parameters as input: \texttt{loss(params, \ldots)}, one can write \texttt{jax.grad(loss, 0)(params, \ldots)}, to get the gradient.
 
\ifarxiv
\section{Example of Lagrangian Forward Model}
\label{sec:example}
To demonstrate how this may work if one has learned
the exact function for $\mathcal{L}$, 
let us study an example of a ball falling
in a gravity $g$ along the direction $q_1$:
\begin{align}
    \mathcal{L} = \frac{1}{2} m \left(\d{q}_1^2+ \d{q}_2^2\right) - m g q_1,
\end{align}
where $g$ is the local scalar gravitational field,
and $m$ is the mass of the ball.
To obtain the dynamics with our forward model,
we calculate the required derivatives which gives us:
\begin{align}
    \nabla_{\dot{q}}\nabla_{\dot{q}}^\top \mathcal{L} &= \left(\begin{array}{cc} m & 0 \\ 0 & m\end{array}\right),\\
        \nabla_q \nabla_{\dot{q}}^\top \mathcal{L} &= 0, \text{ and}\\
    \nabla_{q} \mathcal{L} &= \left(\begin{array}{c} -m g \\ 0\end{array}\right).
\end{align}{}
Thus, we find
\begin{align}
    \left( \begin{array}{c}\dd{q}_1 \\ \dd{q}_2 \end{array} \right)
        &=
    (\nabla_{\dot{q}}\nabla_{\dot{q}}^\top \mathcal{L})^{-1}\left[
    \nabla_q \mathcal{L} - (\nabla_q \nabla_{\dot{q}}^\top \mathcal{L})\dot{q}\right] \\
&= \left(\begin{array}{cc} m & 0 \\ 0 & m\end{array}\right)^{-1}\left[
    \left(\begin{array}{c} -m g \\ 0\end{array}\right)
    \right] \\
    &= \left( \begin{array}{c} -g \\ 0 \end{array} \right)
\end{align}
which simply says that the particle accelerates downwards along $q_1$,
and moves at constant velocity along $q_2$.

\fi 
 
\section{Initialization}
\label{sec:init}
Regular optimization techniques such as Kaiming \citep{kaiming} and Xavier \citep{xavier} initialization are optimized so that in a regular neural network, the gradients of the output with respect to each parameter will have a mean of zero and standard deviation of one. Since the unusual optimization objective of a Lagrangian Neural Network is very nonlinear with respect to its parameters, we found that classical initialization schemes were insufficient.

To find a better initialization scheme, we conducted an empirical optimization of the KL-divergence of the gradient of each parameter with respect to a univariate Gaussian. We repeated this over a variety of neural network depths and widths and fit an empirical formula to our results.

To do this, we ran 2500 optimization steps with different initialization variances on each layer for an MLP of fixed depth and width. Biases were always initialized to zero. We recorded the optimized $\sigma$ values over $\sim$200 random hyperparameter settings with the number of hidden nodes between 50 and 300 and the number of hidden layers between one and three. Then, we used \textit{eureqa} \citep{schmidt2009distilling-free-form-natural}  to find fit an equation using symbolic regression that predicted the optimal initialization variance as a function of the hyperparameters:
\begin{align}
    \sigma = \frac{1}{\sqrt{n}}\left\{
    \begin{array}{cc}
        2.2 & \text{First layer}\\
        0.58i & \text{Hidden layer } i\in\{1,\ldots\}\\
        n, & \text{Output layer,}
    \end{array}
    \right.
\end{align}
This model was optimized for 2 input coordinates and 2 input coordinate velocities which were sampled from univariate Gaussians.

During training, the hidden weight matrices had dimensions ${n\times n}$ and each weight matrix was sampled from $\mathcal{N}(0, \sigma^2)$.
A 100-node 4-layer model would have weight matrices with
shapes $\{(4, 100), (100, 100), (100, 100), (100, 1)\}$ and each one had initializations sampled from
$\{
\mathcal{N}(0, 0.22),
\mathcal{N}(0, 0.058),
\mathcal{N}(0, 0.116),
\mathcal{N}(0, 10)
\},$
respectively.

\end{document}